\documentclass[letterpaper, 10 pt, conference]{ieeeconf}
\usepackage{amsmath}
\usepackage{mathtools}
\usepackage{algorithm}
\usepackage{setspace}
\usepackage[noend]{algpseudocode}
\usepackage{hyperref}
\usepackage{bm}
\usepackage{hhline}
\usepackage{multirow}

\usepackage{siunitx}

\usepackage{subcaption}
\usepackage{kantlipsum} 
\usepackage{mwe} 
\usepackage{booktabs}

\DeclareCaptionLabelSeparator{periodspace}{.\quad}
\IEEEoverridecommandlockouts                              

\title{\LARGE \bf
MACE: Multi-Agent Autonomous Collaborative Exploration of Unknown Environments
}

\author{Charbel Toumieh and Alain Lambert
\thanks{The two authors are with the Universit\'e Paris-Saclay, CNRS, Laboratoire Interdisciplinaire des Sciences du Numérique, 91405, Orsay, France \url{https://www.lisn.upsaclay.fr}}%
}
\begin{document}

\maketitle
\thispagestyle{empty}
\pagestyle{empty}

\begin{abstract}
In this paper, we propose a new framework for multi-agent collaborative exploration of unknown environments. The proposed method combines state-of-the-art algorithms in mapping, safe corridor generation and multi-agent planning. It first takes a volume that we want to explore, then proceeds to give the multiple agents different goals in order to explore a voxel grid of that volume. The exploration ends when all voxels are discovered as free or occupied, or there is no path found for the remaining undiscovered voxels. The state-of-the-art planning algorithm uses time-aware Safe Corridors to guarantee intra-agent collision safety as well safety from static obstacles. The presented approach is tested in a state of the art simulator for up to 4 agents.  
\end{abstract}
\textbf{ }

 \textbf{video}: \url{https://youtu.be/v7P7HpBRY50}

\section{INTRODUCTION}
\subsection{Problem statement}
Multi-agent exploration has numerous real world applications such as search and rescue, infrastructure inspection and cave exploration. An exploration framework that is computationally efficient and safe is thus tremendously beneficial. It is the purpose of this paper to present a new method for multi-agent exploration that is suitable for low compute embedded systems. 

\subsection{Related work}
\subsubsection{Single multirotor planning}
Many works address single quadrotor planning. Recent state-of-the-art contributions use Safe Corridors to account for unknown/unexplored part of the environment and guarantee trajectory safety \cite{tordesillas2020faster}, \cite{toumieh2020planning}. Safe Corridors are a series of overlapping convex polyhedra that cover only free space and are used in many autonomous navigation methods \cite{toumieh2022dyn} \cite{toumieh2022time}. In \cite{toumieh2020planning}, the authors use Safe Corridors in a MPC/MIQP formulation that is computationally efficient and accounts for drag forces. Our work is based on their approach with modifications to account for other moving agents.

\subsubsection{Multi-agent planning}
While some works have addressed multi-agent planning \cite{kamel2017robust}, \cite{zhu2019chance}, \cite{lin2020robust}, \cite{Zhu2019bvc}, \cite{Luis2019dmpc}, \cite{toumieh2020multiagent}, \cite{liu2017planning}, only few works address the problem of low compute and safe multi-agent autonomous exploration while avoiding static obstacles and other exploring agents.
In \cite{zhou2021egoswarm} the authors present an autonomous multi-agent planning solution using only on-board  resources. The presented approach does not address full volume exploration and focuses only on intra-agent collision avoidance as well as static obstacle collision avoidance.

\subsubsection{Exploration}
Single quadrotor exploration and frontier analysis/selection has been extensively studied in the literature. Recent work \cite{9324988} has shown significant improvement over other methods such as \cite{cieslewski2017rapid}, \cite{yamauchi1997frontier}, \cite{bircher2016receding}. The proposed method uses a frontier information structure (FIS) that is maintained incrementally. It is provided to the exploration planner that plans exploration motions in three sequential steps: it first finds global tours that are efficient; then it selects a local set of optimal
viewpoints; finally it generates minimum-time local trajectories. However, it is unclear how to scale this approach to multi-agent planning and whether the computational efficiency would be maintained. It is why we inspired our goal selection method from the Classical method \cite{yamauchi1997frontier} for frontier/goal selection, which scales well for multiple agents.

\begin{figure}
\centering
\includegraphics[trim={2cm 0 0 0},clip,width=1\linewidth]{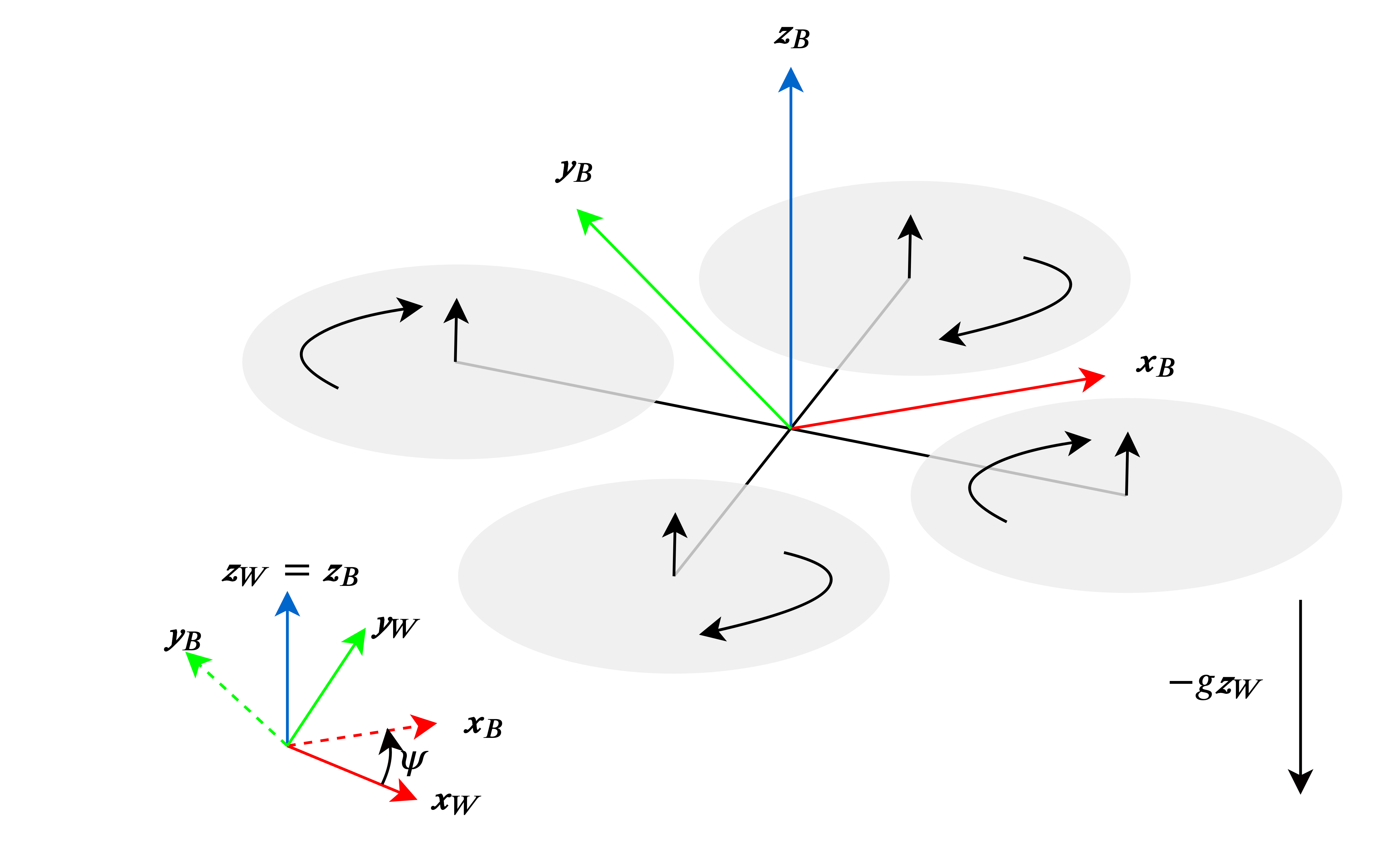}
\caption{Schematics of the multirotor model with the used
coordinate systems.}
\label{fig:quad_fig}
\end{figure}

\subsection{Contribution}
The main contribution of our paper is a novel exploration framework that is based on a decentralized and synchronous planning method. The planning method takes inspiration mainly from \cite{toumieh2020planning} for single quadrotor planning and static obstacle avoidance, \cite{toumieh2020multiagent} for intra-agent collision avoidance, and \cite{yamauchi1997frontier} for goal selection for each agent. The framework also uses recent advances in state-of-the-art mapping \cite{toumieh2020mapping} and Safe Corridor generation \cite{toumieh2020convex} \cite{toumieh2022shapeaware} that makes it low compute and suitable for embedded systems. The planning framework is tested in simulation using the state-of-the-art Airsim \cite{Airsim} simulator.

\subsection{Assumptions}
The following assumptions are made:
\begin{itemize}
    \item The position of each agent is known within a certain range of uncertainty. In the simulation, the position of each agent is known perfectly, but our framework can account for uncertainties by inflating each agents collision radius.
    \item All agents can communicate with each other within a certain delay. The communication only needs to happen when agents are close to each other.
    \item All agents can communicate with a central hub within a certain delay. The central hub will be tasked with merging each agent's map with a global map and sending goals to each agent.
\end{itemize}

\section{Agent Model} \label{sect:model}

\begin{table}[h]
\caption{Nomenclature}
\label{tab:nomenclature}
\begin{center}
\begin{tabular}{c l}
\hline
$g$ & gravity \\
$m$ & multirotor mass \\
$\boldsymbol{p}$ & position vector $x,y,z$ in the world frame\\
$\boldsymbol{v}$ & velocity vector $v_x,v_y,v_z$ in the world frame\\
$\boldsymbol{a}$ & acceleration vector from thrust and gravity in the world frame\\
$\boldsymbol{j}$ & jerk vector $j_x,j_y,j_z$ in the world frame\\
$\boldsymbol{z}_W$ & world frame $z$\\
$\boldsymbol{z}_B$ & body frame $z$\\
$\boldsymbol{R}$ & rotation matrix from body to world frame\\
$\boldsymbol{D}$ & quadratic drag matrix\\
$\phi$ & roll angle\\
$\theta$ & pitch angle\\
$\psi$ & yaw angle\\
$c_{cmd}$ & total thrust command\\
\hline
\end{tabular}
\end{center}
\end{table}

Each agent is modeled after a multirotor whose control inputs are the attitude and thrust. The equations of motion are the following (nomenclature Tab. \ref{tab:nomenclature}):
\begin{align}
     & \Dot{\boldsymbol{p}} = \boldsymbol{v} \\
     & \Dot{\boldsymbol{v}} = -g \boldsymbol{z}_W + \dfrac{c_{cmd}}{m} \boldsymbol{z}_B - \boldsymbol{R}\boldsymbol{D}\boldsymbol{R'}\boldsymbol{v} ||\boldsymbol{v}||_2 \label{eqn:acc}\\
     & \Dot{\phi} = \Dot{\phi}_{cmd} \\
     & \Dot{\theta} = \Dot{\theta}_{cmd} \\
     & \Dot{\psi} = \Dot{\psi}_{cmd} 
\end{align}

The above-mentioned equations are simplified into the following model:

\begin{equation}
  \label{eqn:mot}
    \begin{aligned}
     & \Dot{\boldsymbol{p}} = \boldsymbol{v} \\
     & \Dot{\boldsymbol{v}} = \boldsymbol{a} - \boldsymbol{D}_{lin\_max}\boldsymbol{v} \\
     & \Dot{\boldsymbol{a}} = \boldsymbol{j}
    \end{aligned}
\end{equation}

$\boldsymbol{D}_{lin\_max}$ is a diagonal matrix that represents the maximum linear drag coefficient in all directions. This matrix is identified offline as shown in \cite{toumieh2020planning}. The quadratic drag model is replaced with a linear worst-case scenario model. Many off-the-shelf solvers are very efficient for linear constraints, which is why we transformed the model into a linear one.

\section{The Framework}
\begin{figure}
\centering
\includegraphics[trim={0cm 0 0 -1cm},clip,width=1\linewidth]{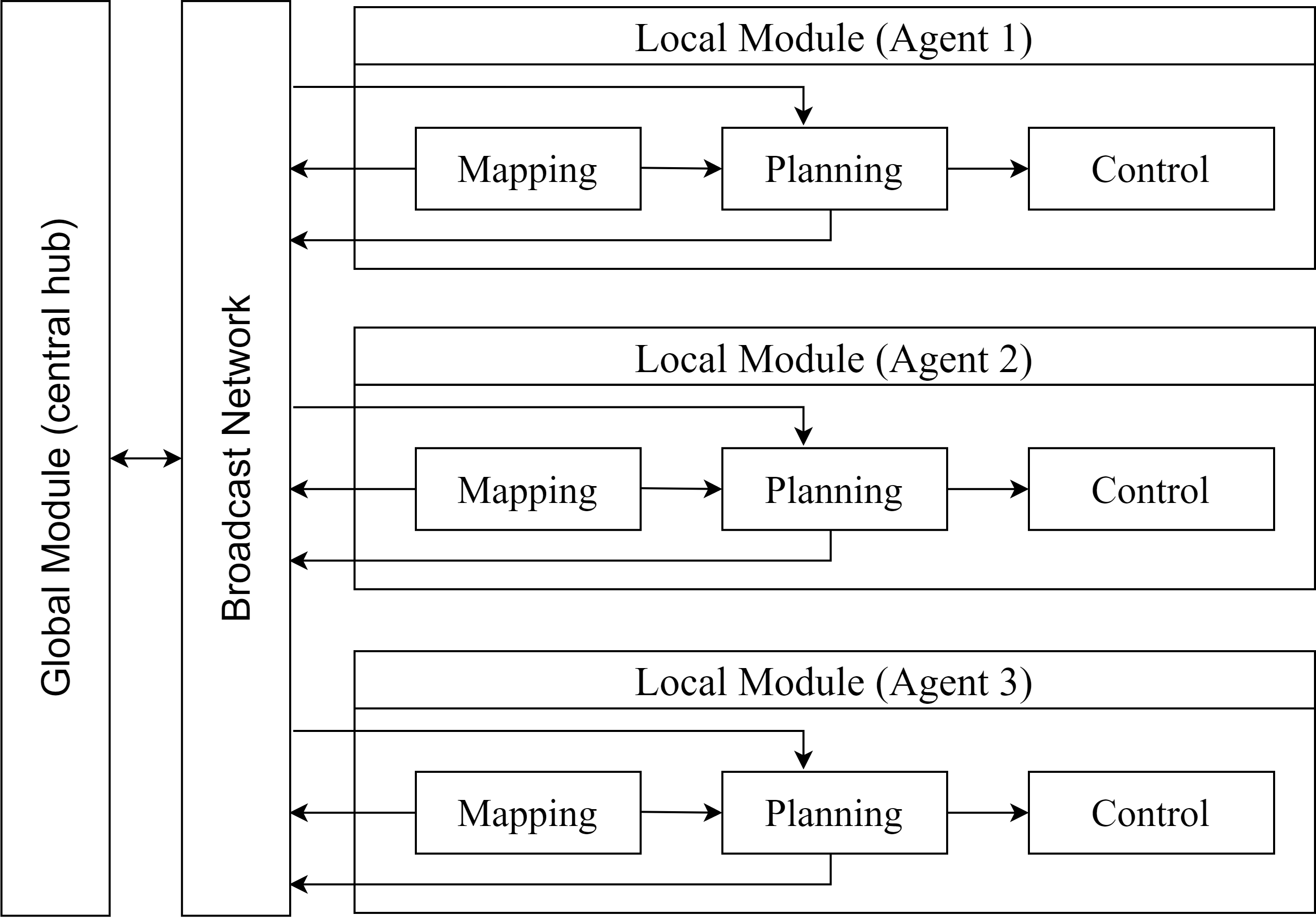}
\caption{We show the global pipeline of the exploration framework with 3 agents. The broadcast network communicates the local maps of each agent to the global module, the goals computed by the global module to the agents, and generated trajectories of all the agents to each agent.}
\label{fig:global_pipeline}
\end{figure}

The framework takes a volume that we want to explore, then proceeds to give all the agents different goals until all the volume is explored. The volume is represented as a voxel grid, and it is considered fully explored when all the voxels are either free or occupied, or no agent can find a path to the remaining unknown voxels.
The framework is divided into 2 modules (Fig. \ref{fig:global_pipeline}):
\begin{enumerate}
    \item A local module that is run on        each agent and that contains     mapping,     planning and control submodules.
    \item A global module that merges all the measurements of all the agents into a single global voxel grid. This module then sends goals to each agent to explore the volume.
\end{enumerate}
Each of these modules as well as each submodule will be detailed separately in this section. We make the following assumptions:
\begin{enumerate}
    \item The positions of all the agents are known within a certain range of uncertainty.
    \item When the agents are close to each other they can communicate their positions and generated trajectories to each others within a certain delay.
    \item The agents can all communicate with a central hub (running the global module) within a certain delay.
\end{enumerate}

\subsection{Local module - run on each agent}
The local module is run on each agent an consists of 3 submodules: mapping, planning and control.
\subsubsection{Mapping}
Each agent is equipped with an omnidirectional lidar that allows is to see in all directions. The output of the lidar is a pointcloud that is transformed into a voxel grid using a GPU accelerated method \cite{toumieh2020mapping}. We remove the points associated to other agents appearing in the lidar scan. This is done by removing all the points of the scan within a certain distance from the center of other agents (which we know due to the communication of trajectories between agents). The local voxel grid moves along with the agent such that the agent's position is always in the center of the grid. Each voxel grid is sent to the global module to be merged into the global voxel grid.

\subsubsection{Planning}
Each agent plans locally to reach the goal sent to it by the global module. The planning framework is the same as \cite{toumieh2020planning}, with modifications made to the Safe Corridor part to become time-aware as presented in \cite{toumieh2020multiagent}. This allows to avoid collision with other agents as well as static obstacles.

\begin{figure}
\centering
\includegraphics[trim={0cm 0 0 -1cm},clip,width=1\linewidth]{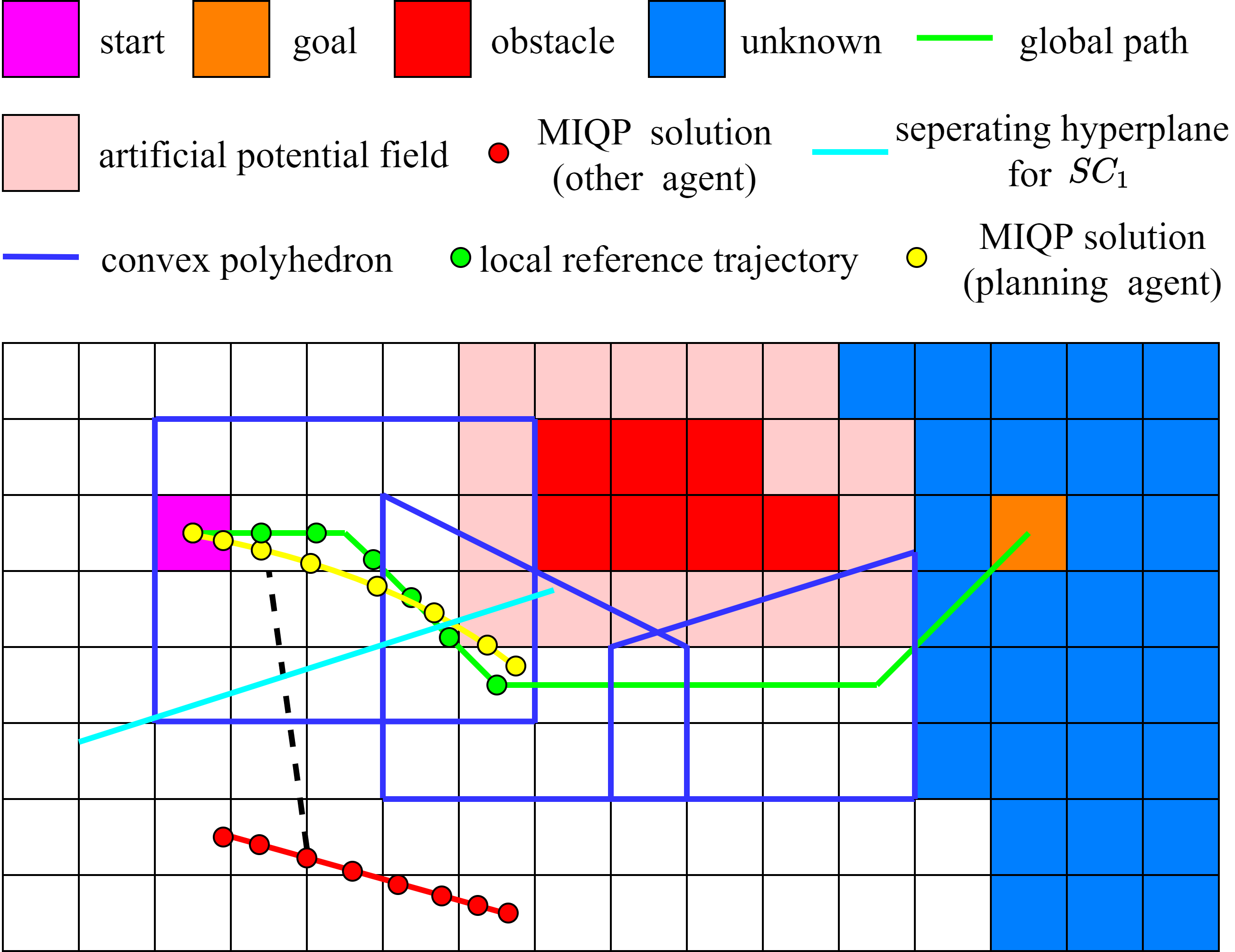}
\caption{We show a planning iteration of our algorithm. The global path is generated using JPS and DMP. DMP uses the artificial potential field (pink) to push the path away from the obstacles. The convex polyhedra (blue) span only the free space. A local reference trajectory is sampled from the global path (green circles). We show the separating hyperplane (cyan) generated between the planning agent and another moving agent using the third predicted position. This hyperplane constraint is added to the convex polyhedron constraints to create $SC_1$ which will be used as constraint for the second and third positions in the next planning iteration. This procedure is done for all positions to create all $SC_i$ (time aware SC). It is then given at the next iteration with the local reference trajectory to the MIQP solver to generate a feasible trajectory (yellow circles).}
\label{fig:resum}
\end{figure}

\begin{figure}
\centering
\includegraphics[trim={0cm 0 0 -1cm},clip,width=1\linewidth]{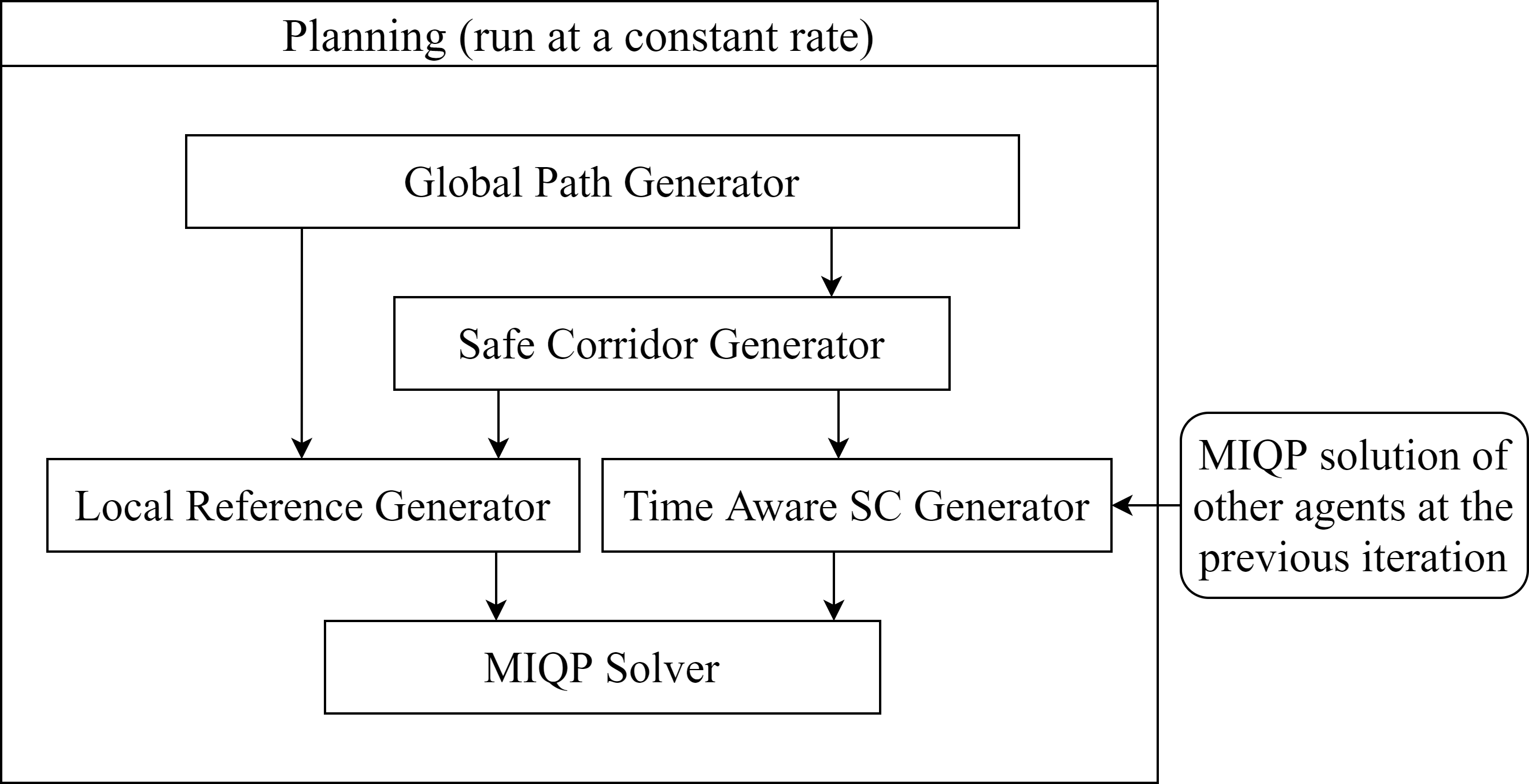}
\caption{We show the global pipeline of the planning framework of a single planning agent.}
\label{fig:diagram}
\end{figure}

The planning framework is divided into the following steps that are executed sequentially and at a constant rate (Fig. \ref{fig:diagram}):
\begin{enumerate}
    \item Generate a global path that avoids static obstacles.
    \item Generate a Safe Corridor.
    \item Generate a time-aware Safe Corridor to avoid collision with other agents.
    \item Generate a local reference trajectory using the global path and the the Safe Corridor.
    \item Solve the MPC/MIQP problem to generate a trajectory close to the local reference trajectory and within the constraints of the time-aware Safe Corridor to guarantee safety.
\end{enumerate}
First a global path is generated using Jump Point Search (JPS) \cite{harabor2011online} and Distance Map Planner (DMP) \cite{jps3d}. JPS is a shortest path algorithm that preserves A*'s optimality, while potentially lowering the computation time by an order of magnitude. The Distance Map Planner (DMP) \cite{jps3d} is used to make the generated path safer. It leverages the artificial potential field to push the path away from the obstacles (Fig. \ref{fig:resum}). We use the local voxel grid to search for the path. If the goal is outside the local voxel grid, an intermediate goal inside it is chosen as shown in \cite{toumieh2020planning}.

Then, we create a Safe Corridor 
around that path to avoid static obstacles using the method described in \cite{toumieh2020convex}. The Safe Corridor covers only free voxels (unknown voxels are considered as obstacles). We only generate $P_{hor}$ (polyhedra horizon used for optimization) overlapping polyhedra around the global path. $P_{hor}$ is typically chosen equal to $2$ to avoid high solving times.

Using the Safe Corridor (SC), we generate $N$ local Safe Corridors $SC_i$, $i=1,2,...,N$ with $N$ the number of  discretization steps of the MPC/MIQP. Each $SC_i$ is generated by adding the separating hyperplanes between the $i^{th}$ predicted position of the agent and the $i^{th}$ predicted position of all the other agents. These positions correspond to the solution of the MPC/MIQP at the last iteration. The generation of the $SC_i$ is detailed in \cite{toumieh2020multiagent}. The series of $SC_i$ is called time-aware SC.

We then generate a local reference trajectory $\boldsymbol{x}_{ref}$ by sampling the global path at a sampling velocity $v_{samp}$ and sampling acceleration $a_{samp}$ starting from the current agent position. These values are a user input but are usually chosen close to the agent dynamical limits. Any sampled point that is outside the SC is replaced by the most recent sampled point that is inside the SC. At each iteration, a new local reference trajectory is generated only if the predicted agent position $\boldsymbol{x}_{N}$ is within $thresh\_dist$ of $\boldsymbol{x}_{N,ref}$. More details about this step can be found in \cite{toumieh2020planning}.

Finally, we use the time-aware SC and the local reference trajectory in a MIQP/MPC formulation to generate a trajectory that is safe from static obstacles and other moving agents. With $\boldsymbol{x} = [\boldsymbol{p} \ \boldsymbol{v} \ \boldsymbol{a}]^T $, $\boldsymbol{u} = \boldsymbol{j}$, $f(\boldsymbol{x}(t),\boldsymbol{u}(t))$ defined by Eq. \ref{eqn:mot}, the model is discretized using Euler to obtain the discrete dynamics $\boldsymbol{x}_{k+1} = f_d(\boldsymbol{x}_k,\boldsymbol{u}_k)$.
The agent velocity is limited by the drag forces. The maximum bounds on the acceleration and the jerk in each direction are determined by the dynamics of the agent.

Collision avoidance is guaranteed by forcing every two consecutive discrete points $k$ and $k+1$ (and thus the segment formed by them) to be in one of the overlapping polyhedra of a given local SC ($SC_k$). Let's assume we have $P$ overlapping polyhedra in $SC_k$. They are described by
$\{(\boldsymbol{A}_{kp}, \boldsymbol{c}_{kp})\}$, $p = 0 : P-1$. The constraint that the discrete position $\boldsymbol{p}_k$ is in a polyhedron $p$ is described by $\boldsymbol{A}_{kp}.\boldsymbol{p}_k \leq \boldsymbol{c}_{kp}$. We introduce binary variables $b_{kp}$ ($P$ variables for each $\boldsymbol{x}_k$, $k = 0 : N-1)$ that indicate that $\boldsymbol{p}_k$ and $\boldsymbol{p}_{k+1}$ are in the polyhedron $p$. We force all the segments to be in at least one of the polyhedra with the constraint $\sum_{p=0}^{P-1} b_{kp} \geq 1$. 

We formulate our MPC under the following Mixed-Integer Quadratic Program (MIQP) formulation. $\boldsymbol{R}_x$, $\boldsymbol{R}_N$ and $\boldsymbol{R}_u$ are the weight matrix for the discrete state errors without the final state, the weight matrix for the final discrete state error (terminal state), and the weight matrix for the input, respectively.

\begin{align}
    & \underset{\substack{\boldsymbol{x}_k,\boldsymbol{u}_k}}{\text{minimize}}
& & \sum_{k=0}^{N-1} (||\boldsymbol{x}_k - \boldsymbol{x}_{k,ref}||_{\boldsymbol{R}_x}^2 +  ||\boldsymbol{u}_k||_{\boldsymbol{R}_u}^2) \nonumber \\
& & & + ||\boldsymbol{x}_N - \boldsymbol{x}_{N,ref}||_{\boldsymbol{R}_N}^2 \label{eq:NLP}\\
& \text{subject to} & & \boldsymbol{x}_{k+1} = f_d(\boldsymbol{x}_k,\boldsymbol{u}_k),\quad k = 0:N-1 \\
& & & \boldsymbol{x}_0 = \boldsymbol{X}_0  \\
& & & \boldsymbol{v}_N = \boldsymbol{0}  \\
& & & \boldsymbol{a}_N = \boldsymbol{0}  \\
& & & |a_{x,k}| \leq a_{x,max} \\
& & & |a_{y,k}| \leq a_{y,max}, \quad a_{z,k} \leq a_{z,max} \\
& & & a_{z,k} \geq a_{z,min}, \quad |j_{x,k}| \leq j_{x,max} \\
& & & |j_{y,k}| \leq j_{y,min}, \quad |j_{z,k}| \leq j_{z,max} \\
& & & b_{kp} = 1 \implies \begin{cases} \boldsymbol{A}_{kp}\boldsymbol{p}_k \leq \boldsymbol{c}_{kp} \\
\boldsymbol{A}_{kp}\boldsymbol{p}_{k+1} \leq \boldsymbol{c}_{kp} \label{eqn:const_poly}
\end{cases}\\
& & & \sum_{p=0}^{P_{hor}-1} b_{kp} \geq 1 \\
& & & b_{kp} \in \{0,1\}
\end{align}

The MIQP is solved using the Gurobi solver \cite{gurobi}.

\subsubsection{Control}
Each agent is controlled using a nonlinear MPC \cite{kamel2017linear}, with the \textbf{acados} toolkit \cite{Verschueren2018}. The MPC minimizes the cost function:
\begin{equation}
\begin{aligned}
    J = & \int_{t=0}^{T}||\boldsymbol{x}(t) - \boldsymbol{x}_{ref}(t)||_{\boldsymbol{Q}_x}^2 + ||\boldsymbol{u}(t) - \boldsymbol{u}_{ref}(t)||_{\boldsymbol{R}_u}^2 dt \\ 
    & + ||\boldsymbol{x}(T) - \boldsymbol{x}_{ref}(T)||_{\boldsymbol{P}}^2
\end{aligned}
\end{equation}

We use the model described in section \ref{sect:model} with $\boldsymbol{u} = [c_{cmd} \ \Dot{\phi}_{cmd} \ \Dot{\theta}_{cmd} \ \Dot{\psi}_{cmd}]^T$ and $\boldsymbol{x} = [\boldsymbol{p} \ \boldsymbol{v} \ \phi \ \theta \ \psi]^T$.

\subsection{Global module - run on a central hub}
This module takes a volume that we want to explore, merges the measurements/maps of all the agents, and sends a goal to each agent at a constant rate until that volume is fully explored. The volume is represented as a voxel grid and is considered fully explored when all the voxels are either free or occupied, or no agent can find a path to the remaining unknown voxels.

\subsubsection{Merging maps from all agents}
Each agent has its local voxel grid that is used for local obstacle avoidance and navigation. When this map is transmitted to the global module, we are able to efficiently merge it into the global map if the voxels of the local and global maps overlap with each others i.e. the vector formed by the origin of the local map and the origin of the global map has its $x$, $y$ and $z$ components as multiples of the voxel size. This can be achieved during the initialization phase, by setting the initial origin of all maps (local and global) to $(0, 0, 0)$. As presented in \cite{toumieh2020mapping}, the origin of the local voxel grid will be always offset in all directions by a multiple of the voxel size to keep the robot at its center.

We only update a voxel of the global grid if its corresponding voxel in the local voxel grid is free or occupied. In the case of static environments, we can add an additional condition to only update a voxel grid in a global grid if its not occupied (and its corresponding voxel in the local voxel grid is free or occupied).

\subsubsection{Computing goals for agents}
\begin{figure}
\centering
\includegraphics[trim={0.4cm 0.4cm 0.5cm 0.03cm},clip,width=1\linewidth]{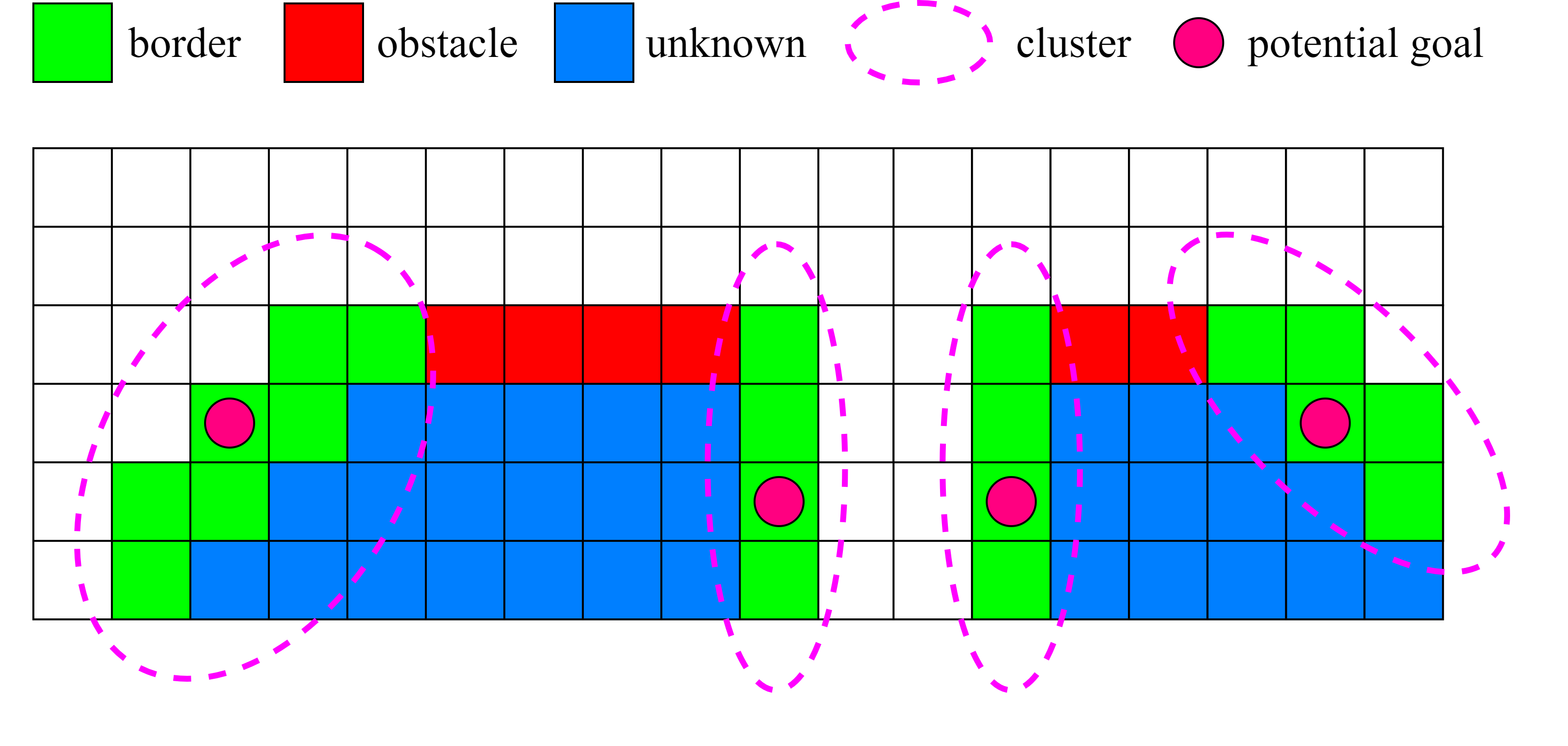}
\caption{We show an example of the borders, clusters and computed potential goals of a voxel grid.}
\label{fig:global_mapper}
\end{figure}

After updating the global map with the local voxel grids of all the agents, we proceed to compute goals for each agent. We define a neighbour voxel as a voxel that is reachable by moving at most 1 voxel unit in each direction from the current voxel. The goals are computed by doing the following steps sequentially (Fig. \ref{fig:global_mapper}):
\begin{enumerate}
    \item Find the border voxels: this is done by going over all the voxels of the global map. If a voxel is free and has a neighbour voxel that is unknown, then it is designated as a border voxel.
    \item Cluster the border voxels: we form clusters out of the border voxels using the following rule: if two border voxel are neighbors, they belong to the same cluster. 
    \item Compute cluster centroid: we compute the centroid of each cluster by averaging the positions of all the voxels that belong to that cluster.
    \item Compute potential goals: we compute potential goals by finding the voxel border in a cluster that is the closest to its centroid. If multiple voxels have the same closest distance, we chose one randomly.
    \item Compute goals: the finals goals are computed by first choosing the closest potential goal to the first agent, and removing it from the potential goals' list so that two agents don't get the same goal. From the remaining goals, we chose the closest one to the second agent and remove it from the list. We continue in this fashion until all the agents have goals, or no potential goals exist anymore.
\end{enumerate}

\begin{table*}[ht]
\centering
\begin{tabular}{c | ccccc}\hline
 Number of agents & Flight distance (m) & Flight velocity (m/s) & Exploration time (s) & Computation time (ms) & Safety ratio \\ \hhline{======}
1 & 239.7 / 263.3 / 20.5 & 1.74 / 3.4 / 0.7 & 139.92 / 158 / 15.6 &  9 / 45 / 3.7 & - \\ \hline
2 & 129.5 / 137.7 / 7.72 & 1.8 / 3.47 / 0.74 & 78.2 / 83.3 / 4.67 & 12 / 49.8 / 4.7 & 1.12 \\ \hline
3 & 81.2 / 110.91 / 13.4 & 1.9 / 3.62 / 0.81 & 49.6 / 60.2 / 6.36 & 13 / 61 / 5.2 & 1.03 \\ \hline
4 & 73 / 86.8 / 9.2 & 2.04 / 3.9 / 0.88 & 44.9 / 50 / 4.27 & 14.5 / 64 / 5.3 & 1.13 
\end{tabular}
\caption{Comparison between 1,2,3 and 4 agents using our framework on 5 randomly generated maps of size $30 \ m \times 30\ m\times 3\ m$ and with obstacle density $0.1 \ obst/m^2$. We show the \textbf{mean / max / standard deviation} per agent of each metric, except for the Safety ratio.}
\label{table:comparison_table}
\end{table*}

\begin{figure*}
\begin{subfigure}{0.5\textwidth}
\centering
\includegraphics[trim={1cm 0cm 1cm 0cm},clip,width=0.95\linewidth]{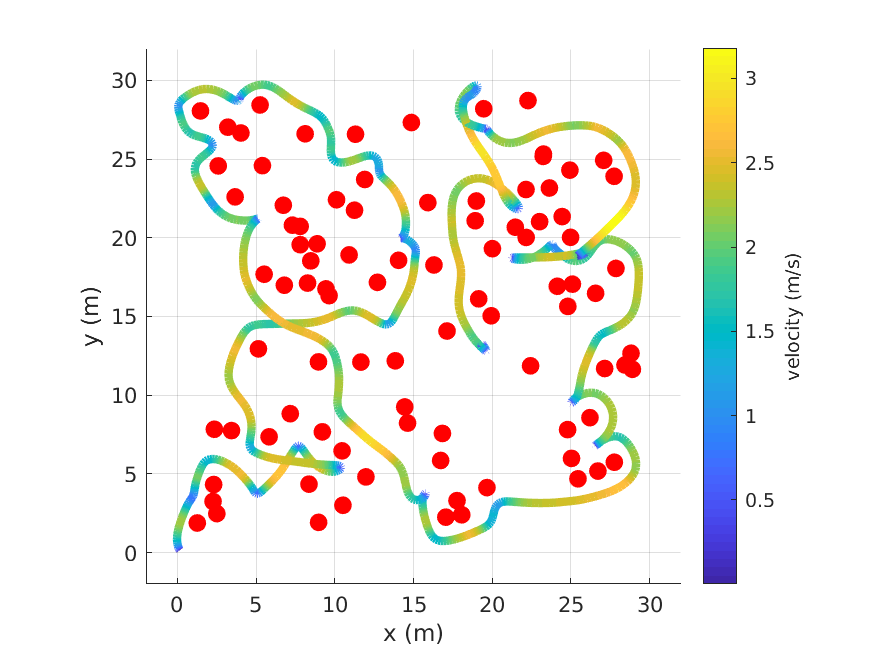}
\caption{Exploration with 1 agent}
\label{fig:comparison_1}
\end{subfigure}
\begin{subfigure}{0.5\textwidth}
\centering
\includegraphics[trim={1cm 0cm 1cm 0cm},clip,width=0.95\linewidth]{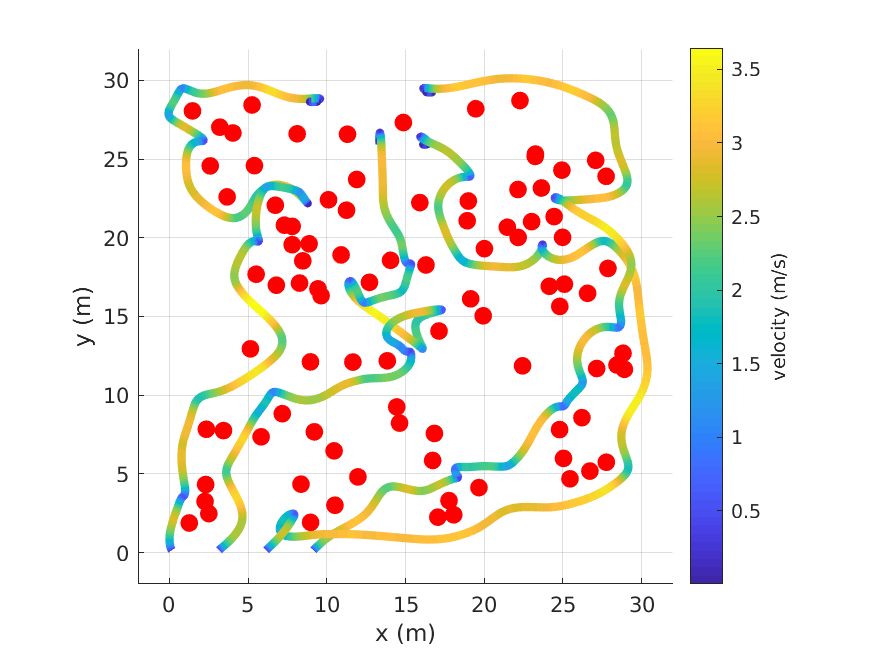}
\caption{Exploration with 4 agents}
\label{fig:comparison_4}
\end{subfigure}

\caption{The trajectories and velocity of 1 exploring agent and 4 exploring agents on the same map. The initial position of agents are offset by $3\ m$ from each others in the $x$ direction in the case of 4 agents}
\label{fig:comparison}
\end{figure*}

\section{Simulation}
The simulation is done in a $30\ m\times30\ m\times3\ m$ environment that contains 90 cylinder obstacles of radius $0.35\ m$ and height $3 \ m$. Their positions are generated randomly, following a uniform distribution. The environment is generated in Airsim \cite{Airsim}.
The Gurobi solver is set to use one thread only as this resulted in faster computation times during our simulations.
All testing is done on the Intel Core i7-9750H up to 4.50 GHz CPU and NVIDIA's GeForce RTX 2060 up to 1.62 GHz. We simulate up to 4 agents due to the limited computing power to run the Airsim simulation, mapping/planning/control for each agent and the global module. The agents communicate between themselves using a ROS \cite{ros} topic. They also communicate with the global module 
using a ROS topic.

\subsection{Mapping parameters}
The local map of each agent is of size $20\ m\times30\ m\times3\ m$. The global map size is $30\ m\times30\ m\times3\ m$. The voxel size used for both maps is $0.3\ m$. The update rate for the local map is $10$ Hz (lidar measurement frequency) and the update rate of the global map is $5$ Hz. Every time the global map is updated, a new set of goals is computed and sent to each agent. The local map update takes less than $1\ ms$ due to the GPU accelerated method we use \cite{toumieh2020mapping} and the global map update and goal computation takes less than $3\ ms$.

\subsection{Planner parameters}
We choose the following parameters: $N = 12$, $h = 100 \ ms$, $a_{samp} = a_{x,max} = a_{y,max} = a_{z,max} = 0.7*g$, $a_{z,min} = -g$, $j_{x,max} = j_{y,max} = j_{z,max} = 8 \ m/s^2$, $v_{samp} = 3.5 \ m/s$, $\boldsymbol{D}_{lin\_max} = diag(1,1,1)$, $thresh\_dist = 0.4 \ m$, $P_{hor} = 2$. The weight matrices are diagonal: $\boldsymbol{R}_x = diag(200,200,200,0,0,0,0,0,0)$, $\boldsymbol{R}_N = diag(100,100,100,0,0,0,0,0,0)$ and $\boldsymbol{R}_u = diag(0.01,0.01,0.01)$. The drone radius is $d_{rad} = 0.3 \ m$. The DMP planner pushes the JPS path $0.6\ m$ (2 voxels) away from obstacles (when possible). The planning frequency is $10$ Hz. The planning requires synchronized clocks between the agents. This is satisfied as all the agents use the same CPU clock during the simulation. In a real world situation, the clocks can be synchronized at the beginning of the exploration since the drift is minimal (1 microsecond every second) for the duration of an exploration task. 

\subsection{Controller parameters}
The frequency of the MPC controller is $20$ Hz. The weights of the controller are:
\begin{align}
    \boldsymbol{P} = \boldsymbol{Q}_x = diag(15,15,15,0.01,0.01,0.01,0,0,1) \\
    \boldsymbol{R}_u = diag(0.05,0.1,0.1,0.1)
\end{align}
All parameters are set by approximation/experimentation and may not be optimal. We limit $|\phi| \leq 85 \ deg$, $\theta \leq 85 \ deg$, $|\Dot{\phi}_{cmd}| \leq 120 \ deg/s$, $|\Dot{\theta}_{cmd}| \leq 120 \ deg/s$ and $|\Dot{\psi}_{cmd}| \leq 60 \ deg/s$.

\subsection{Simulation results}

We show the results of doing a series of 5 explorations on 5 randomly generated maps using 1,2,3 and 4 agents in Tab. \ref{table:comparison_table}. We also show the performance of 1 agent and 4 agents on the same map (Fig. \ref{fig:comparison}). The starting position of the first agent is $(0,0,0)\ m$. When using multiple agents, the starting position of agents are offset from each others $3\ m$ in the $x$ direction (Fig. \ref{fig:comparison_4}). All agents stop moving when the exploration is done i.e. all voxels are discovered as free or occupied, or there is no path found for the remaining undiscovered voxels.

The average flight distance per agent decreases in a way that is approximately inversely proportional to the number of agents used. The average flight velocity slightly increases since an agent has less chances of being \textit{torn} between two close goals as other agents explore goals that are close to it. The average exploration time which corresponds to the time required to explore the full volume also decreases in an approximately inversely proportional way to the number of agents used. The average computation time increases as we use more agents due to the fact that there are more constraints that are added to each agent's MIQP/MPC that correspond to the separating hyperplanes with other agents. Note that the computation time can be decreased by lowering the number of discretization steps $N$. Since the maximum computation time is $64\ ms$, this would leave $100 - 64 = 36\ ms$ to communicate the trajectory to other agents using a broadcasting network (in our case a ROS topic). This means that we can modify $N$ to lower the computation time enough to satisfy communication constraints. Finally we show the safety ratio which corresponds to the ratio: (minimum distance between any two agents at any time during the exploration)/(minimum collision distance). The planner satisfies safety constraints when using any number of agents.

\section{Conclusions and Future Works}
In this paper, we presented a novel framework for multi-agent autonomous collaborative exploration of unknown environments. The method uses recent advances in Safe Corridor generation, high-speed mapping and multi-agent planning to guarantee safety and computational efficiency during the exploration task. The framework is tested in a state-of-the-art simulator using up to 4 agents.

In the future, we plan to implement the exploration framework in a fully autonomous setting. This requires to add one more submodule in the local module for localization/odometry and account for localization uncertainty in the planning submodule to guarantee safety. 






\bibliographystyle{IEEEtran}
\bibliography{IEEEabrv,IEEEexample}

\begin{thebibliography}{10}
\providecommand{\url}[1]{#1}
\csname url@rmstyle\endcsname
\providecommand{\newblock}{\relax}
\providecommand{\bibinfo}[2]{#2}
\providecommand\BIBentrySTDinterwordspacing{\spaceskip=0pt\relax}
\providecommand\BIBentryALTinterwordstretchfactor{4}
\providecommand\BIBentryALTinterwordspacing{\spaceskip=\fontdimen2\font plus
\BIBentryALTinterwordstretchfactor\fontdimen3\font minus
  \fontdimen4\font\relax}
\providecommand\BIBforeignlanguage[2]{{%
\expandafter\ifx\csname l@#1\endcsname\relax
\typeout{** WARNING: IEEEtran.bst: No hyphenation pattern has been}%
\typeout{** loaded for the language `#1'. Using the pattern for}%
\typeout{** the default language instead.}%
\else
\language=\csname l@#1\endcsname
\fi
#2}}

\bibitem{tordesillas2020faster}
J.~Tordesillas, B.~T. Lopez, M.~Everett, and J.~P. How, ``Faster: Fast and safe
  trajectory planner for flights in unknown environments,'' \emph{arXiv
  preprint arXiv:2001.04420}, 2020.

\bibitem{toumieh2020planning}
C.~Toumieh and A.~Lambert, ``High-speed planning in unknown environments for
  multirotors considering drag,'' in \emph{2021 IEEE International Conference
  on Robotics and Automation (ICRA)}, 2021, pp. 7844--7850.

\bibitem{toumieh2022dyn}
\BIBentryALTinterwordspacing
------, ``Multirotor planning in dynamic environments using temporal safe
  corridors,'' 2022. [Online]. Available:
  \url{https://arxiv.org/abs/2208.06950}
\BIBentrySTDinterwordspacing

\bibitem{toumieh2022time}
------, ``Near time-optimal trajectory generation for multirotors using
  numerical optimization and safe corridors,'' \emph{Journal of Intelligent \&
  Robotic Systems}, vol. 105, no.~1, pp. 1--10, 2022.

\bibitem{kamel2017robust}
M.~Kamel, J.~Alonso-Mora, R.~Siegwart, and J.~Nieto, ``Robust collision
  avoidance for multiple micro aerial vehicles using nonlinear model predictive
  control,'' in \emph{2017 IEEE/RSJ International Conference on Intelligent
  Robots and Systems (IROS)}.\hskip 1em plus 0.5em minus 0.4em\relax IEEE,
  2017, pp. 236--243.

\bibitem{zhu2019chance}
H.~Zhu and J.~Alonso-Mora, ``Chance-constrained collision avoidance for mavs in
  dynamic environments,'' \emph{IEEE Robotics and Automation Letters}, vol.~4,
  no.~2, pp. 776--783, 2019.

\bibitem{lin2020robust}
J.~Lin, H.~Zhu, and J.~Alonso-Mora, ``Robust vision-based obstacle avoidance
  for micro aerial vehicles in dynamic environments,'' 2020.

\bibitem{Zhu2019bvc}
H.~{Zhu} and J.~{Alonso-Mora}, ``B-uavc: Buffered uncertainty-aware voronoi
  cells for probabilistic multi-robot collision avoidance,'' in \emph{2019
  International Symposium on Multi-Robot and Multi-Agent Systems (MRS)}, 2019,
  pp. 162--168.

\bibitem{Luis2019dmpc}
C.~E. {Luis}, M.~{Vukosavljev}, and A.~P. {Schoellig}, ``Online trajectory
  generation with distributed model predictive control for multi-robot motion
  planning,'' \emph{IEEE Robotics and Automation Letters}, vol.~5, no.~2, pp.
  604--611, 2020.

\bibitem{toumieh2020multiagent}
C.~Toumieh and A.~Lambert, ``Decentralized multi-agent planning using model
  predictive control and time-aware safe corridors,'' \emph{IEEE Robotics and
  Automation Letters}, pp. 1--8, 2022.

\bibitem{liu2017planning}
S.~Liu, M.~Watterson, K.~Mohta, K.~Sun, S.~Bhattacharya, C.~J. Taylor, and
  V.~Kumar, ``Planning dynamically feasible trajectories for quadrotors using
  safe flight corridors in 3-d complex environments,'' \emph{IEEE Robotics and
  Automation Letters}, vol.~2, no.~3, pp. 1688--1695, 2017.

\bibitem{zhou2021egoswarm}
X.~Zhou, J.~Zhu, H.~Zhou, C.~Xu, and F.~Gao, ``Ego-swarm: A fully autonomous
  and decentralized quadrotor swarm system in cluttered environments,'' 2021.

\bibitem{9324988}
B.~Zhou, Y.~Zhang, X.~Chen, and S.~Shen, ``Fuel: Fast uav exploration using
  incremental frontier structure and hierarchical planning,'' \emph{IEEE
  Robotics and Automation Letters}, vol.~6, no.~2, pp. 779--786, 2021.

\bibitem{cieslewski2017rapid}
T.~Cieslewski, E.~Kaufmann, and D.~Scaramuzza, ``Rapid exploration with
  multi-rotors: A frontier selection method for high speed flight,'' in
  \emph{2017 IEEE/RSJ International Conference on Intelligent Robots and
  Systems (IROS)}.\hskip 1em plus 0.5em minus 0.4em\relax IEEE, 2017, pp.
  2135--2142.

\bibitem{yamauchi1997frontier}
B.~Yamauchi, ``A frontier-based approach for autonomous exploration,'' in
  \emph{Proceedings 1997 IEEE International Symposium on Computational
  Intelligence in Robotics and Automation CIRA'97.'Towards New Computational
  Principles for Robotics and Automation'}.\hskip 1em plus 0.5em minus
  0.4em\relax IEEE, 1997, pp. 146--151.

\bibitem{bircher2016receding}
A.~Bircher, M.~Kamel, K.~Alexis, H.~Oleynikova, and R.~Siegwart, ``Receding
  horizon" next-best-view" planner for 3d exploration,'' in \emph{2016 IEEE
  international conference on robotics and automation (ICRA)}.\hskip 1em plus
  0.5em minus 0.4em\relax IEEE, 2016, pp. 1462--1468.

\bibitem{toumieh2020mapping}
\BIBentryALTinterwordspacing
C.~Toumieh and A.~Lambert, ``Gpu accelerated voxel grid generation for fast mav
  exploration,'' 2021. [Online]. Available:
  \url{https://arxiv.org/abs/2112.13169}
\BIBentrySTDinterwordspacing

\bibitem{toumieh2020convex}
------, ``Voxel-grid based convex decomposition of 3d space for safe corridor
  generation,'' \emph{Journal of Intelligent \& Robotic Systems}, vol. 105,
  no.~4, pp. 1--13, 2022.

\bibitem{toumieh2022shapeaware}
------, ``Shape-aware safe corridors generation using voxel grids,'' 2022.

\bibitem{Airsim}
\BIBentryALTinterwordspacing
S.~Shah, D.~Dey, C.~Lovett, and A.~Kapoor, ``Airsim: High-fidelity visual and
  physical simulation for autonomous vehicles,'' in \emph{Field and Service
  Robotics}, 2017. [Online]. Available: \url{https://arxiv.org/abs/1705.05065}
\BIBentrySTDinterwordspacing

\bibitem{harabor2011online}
D.~D. Harabor and A.~Grastien, ``Online graph pruning for pathfinding on grid
  maps,'' in \emph{Twenty-Fifth AAAI Conference on Artificial Intelligence},
  2011.

\bibitem{jps3d}
K.~Robotics, ``Mrsl jump point search planning library v1.1,''
  \url{https://github.com/KumarRobotics/jps3d}, accessed 2020-05-09.

\bibitem{gurobi}
\BIBentryALTinterwordspacing
L.~Gurobi~Optimization, ``Gurobi optimizer reference manual,'' 2020. [Online].
  Available: \url{http://www.gurobi.com}
\BIBentrySTDinterwordspacing

\bibitem{kamel2017linear}
M.~Kamel, M.~Burri, and R.~Siegwart, ``Linear vs nonlinear mpc for trajectory
  tracking applied to rotary wing micro aerial vehicles,''
  \emph{IFAC-PapersOnLine}, vol.~50, no.~1, pp. 3463--3469, 2017.

\bibitem{Verschueren2018}
R.~Verschueren, G.~Frison, D.~Kouzoupis, N.~van Duijkeren, A.~Zanelli,
  R.~Quirynen, and M.~Diehl, ``Towards a modular software package for embedded
  optimization,'' in \emph{Proceedings of the IFAC Conference on Nonlinear
  Model Predictive Control (NMPC)}, 2018.

\bibitem{ros}
\BIBentryALTinterwordspacing
{Stanford Artificial Intelligence Laboratory et al.}, ``Robotic operating
  system.'' [Online]. Available: \url{https://www.ros.org}
\BIBentrySTDinterwordspacing

\end{thebibliography}

\end{document}